\documentclass[10pt,twocolumn,letterpaper,dvipsnames]{article}

\usepackage{svg}
\usepackage[pagenumbers]{iccv} %

\newcommand{\bI}{\mathbf{I}}

\newcommand{\bw}{\mathbf{w}}

\newcommand{\bv}{\mathbf{v}}

\newcommand{\bff}{\mathbf{f}}

\newcommand{\bc}{\mathbf{c}}
\newcommand{\bm}{\mathbf{m}}

\newcommand{\ba}{\mathbf{a}}

\newcommand{\cL}{\mathcal{L}}

\newcommand{\figref}[1]{\Fig~\ref{#1}}
\newcommand{\secref}[1]{Section~\ref{#1}}

\newcommand{\tabref}[1]{Table~\ref{#1}}

\makeatletter
\DeclareRobustCommand\onedot{\futurelet\@let@token\@onedot}
\def\@onedot{\ifx\@let@token.\else.\null\fi\xspace}
\def\eg{e.g\onedot} 
\def\ie{i.e\onedot} 
 
\def\etc{etc\onedot} \def\vs{vs\onedot}

\def\Fig{Fig\onedot}   
\makeatother

\newcommand{\xdownarrow}[1]{%
  {\left\downarrow\vbox to #1{}\right.\kern-\nulldelimiterspace}
}

\newcommand{\xuparrow}[1]{%
  {\left\uparrow\vbox to #1{}\right.\kern-\nulldelimiterspace}
}

\newcommand{\boldparagraph}[1]{\vspace{0.15cm}\noindent{\bf #1:} }

\definecolor{First}{HTML}{BDE6CD}%
\definecolor{Second}{HTML}{E2EEBC}%
\definecolor{Third}{HTML}{FFF8C5}%

\def\lm{large models }
\def\vtwo{Leaderboard-v2 }
\def\vone{Leaderboard-v1 }

\definecolor{iccvblue}{rgb}{0.21,0.49,0.74}
\usepackage[pagebackref,breaklinks,colorlinks,allcolors=iccvblue,linkcolor=red]{hyperref}

\title{ETA: Efficiency through Thinking Ahead,\\A Dual Approach to Self-Driving with Large Models} 

\author{
Shadi Hamdan$^{1,2\ast}$\quad Chonghao Sima$^{3}$\quad Zetong Yang$^{4}$\quad Hongyang Li$^{3}$\quad Fatma G\"{u}ney$^{1,2}$
\\[2mm]
$^1$Ko\c{c} University
\quad
$^2$KUIS AI Center
\quad
$^3$The University of Hong Kong
\quad
$^4$OpenDriveLab 
\\[2mm]
\url{https://github.com/OpenDriveLab/ETA}
}

\begin{document}
\maketitle

{\let\thefootnote \relax \footnote{$^\ast$Work partially done during internship at OpenDriveLab. \\
Primary contact:
\texttt{shamdan17@ku.edu.tr}}}

\begin{abstract}
How can we benefit from large models without sacrificing inference speed, a common dilemma in self-driving systems?  A prevalent solution is a dual-system architecture, employing a small model for rapid, reactive decisions and a larger model for slower but more informative analyses. Existing dual-system designs often implement parallel architectures where inference is either directly conducted using the large model at each current frame or retrieved from previously stored inference results. However, these works still struggle to enable large models for a timely response to every online frame. Our key insight is to shift intensive computations of the current frame to previous time steps and perform a batch inference of multiple time steps to make large models respond promptly to each time step. To achieve the shifting, we introduce Efficiency through Thinking Ahead (ETA), an asynchronous system designed to: (1) propagate informative features from the past to the current frame using future predictions from the large model, (2) extract current frame features using a small model for real-time responsiveness, and (3) integrate these dual features via an action mask mechanism that emphasizes action-critical image regions. Evaluated on the Bench2Drive CARLA Leaderboard-v2 benchmark, ETA advances state-of-the-art performance by 8\% with a driving score of 69.53 while maintaining a near-real-time inference speed at 50 ms. Code and checkpoints can be found \href{https://github.com/OpenDriveLab/ETA}{here.}
\end{abstract}    
\section{Introduction}
\label{sec:intro}

\begin{figure}[t]
    \centering
    \includegraphics[width=\linewidth]{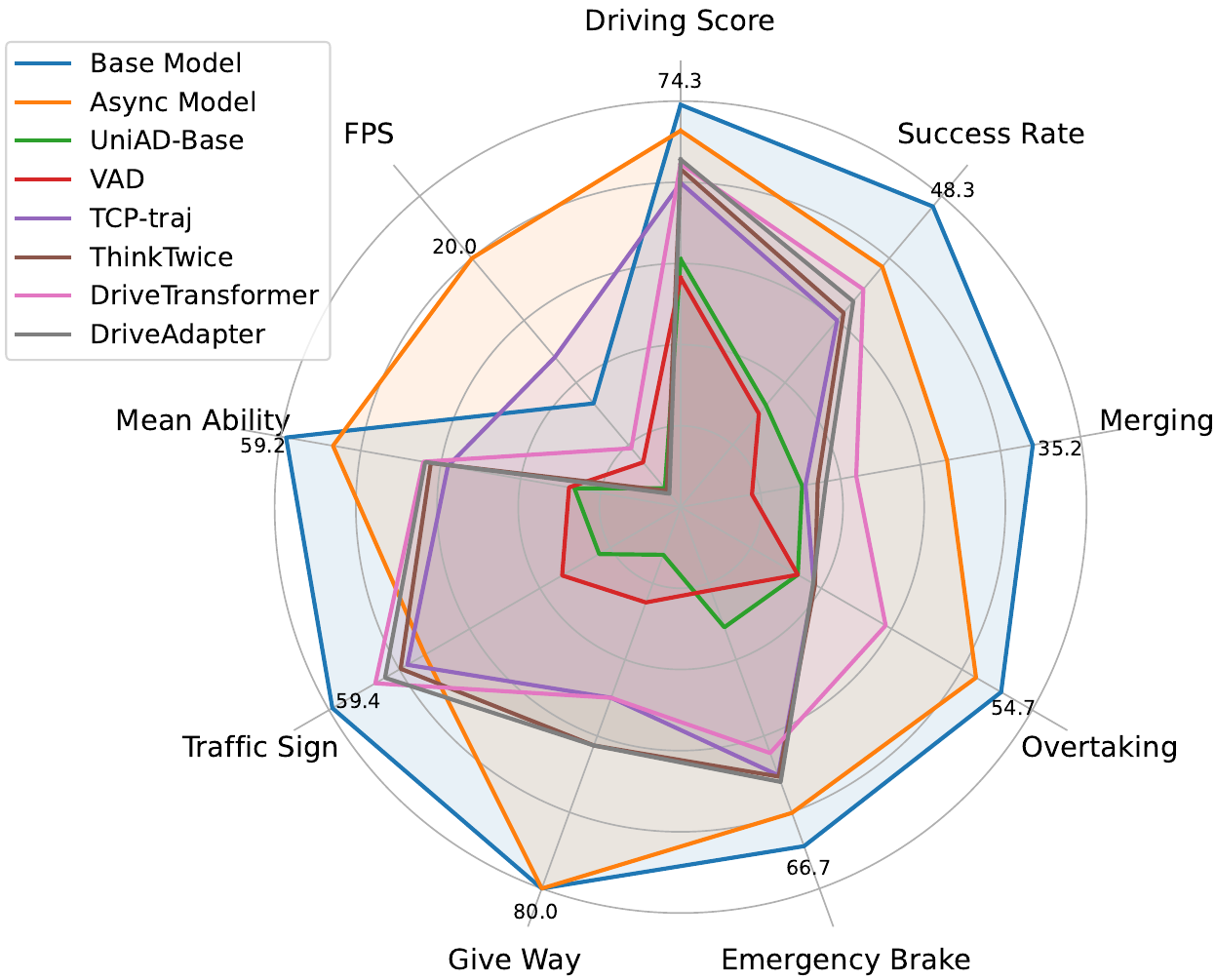}
    \caption{\textbf{Comparison on Bench2Drive~\cite{Jia2024NeurIPS}.} 
    Our \textcolor{RoyalBlue}{Base model} achieves the best performance across all metrics with a high latency. By improving latency to 20 FPS, our \textcolor{orange}{Async model} achieves the second best in all metrics except Traffic Sign handling.}
    \label{fig:ds_vs_fps}
\end{figure}

Large models, including multimodal foundation models~\cite{chatgpt,Chen2024CVPR} and vision foundation models with huge amount of parameters~\cite{Lee2019CVPRWORK,Dosovitskiy2020ICLR}, have been increasingly used in self-driving because of their impressive perception and reasoning capabilities~\cite{Mao2023ARXIV,Xu2024RAL,chen2024PAMI,Hu2023CVPR}. Due to the real-time demand of self-driving, inference-time speed becomes a critical concern when using large models. For instance, the best-performing models \cite{Jia2025ICLR,Jia2023ICCV} on the Bench2Drive closed-loop benchmark \cite{Jia2024NeurIPS} currently have latencies in the order of hundreds of milliseconds due to their large transformer structures, which do not satisfy the real-time requirement. In this work, we address the challenge of utilizing \lm for self-driving \textit{without} compromising inference-time speed. 

The efficiency bottleneck in \lm has driven researchers to develop dual approaches for embodied systems~\cite{Bu2024ARXIV, Li2024ARXIV, Zhang2024CoRL, Yu2024ARXIV, Shi2025ARXIV, Figure2025}, including self-driving~\cite{Tian2024CoRL, Zhang2024ARXIV, Mei2024NeurIPS,sima2025centaur}. Inspired by the System-1 and System-2 framework from cognitive science~\cite{Kahneman2011}, these dual frameworks utilize a small model for fast, reactive decisions and a large model for slower, more thoughtful decisions.
While the dual approach facilitates the use of large models in online systems, the inference results from these models in existing paradigms~\cite{Tian2024CoRL, Zhang2024ARXIV} are often infrequently available. 
Most outcomes are derived from immediate small model inference coupled with delayed large model inference (See \figref{fig:dual_comp}). 
One potential solution for integrating large models into every frame is to store the outputs of the large model for complex cases in memory and retrieve relevant instances as needed using the small model, as explored in prior research~\cite{Mei2024NeurIPS}. However, managing an expanding memory bank in dynamic environments poses a challenge and can be difficult to generalize. In this paper, we manage to make the inference of \lm timely available for each frame.

Our key insight is to enable large models for online inferencing every frame by an {\bf E}fficiency through {\bf T}hinking {\bf A}head (ETA) pipeline, shifting the time-consuming computations of large models on the current frame to a previous time step, and performing a batch inference for multiple frames on one time to trade space complexity for time complexity, as shown in \cref{fig:dual_comp} (right).
Consequently, each online frame is processed using large models from prior time steps, yielding outcomes that integrate both timely large model inference and small model inference.

However, shifting the computation of the current time step to a prior time step is non-trivial. As the information of the current time step is not available in previous frames.
Our second novelty lies in updating the information from the prior time step for invisible online time step through
i) propagating accurate information from the large model via future prediction; ii) processing the current time step with a small model to capture changes that are difficult to predict. %
Furthermore, we encourage observations and predicted actions to align more closely in the feature space, enabling the dual framework to function effectively. 
We design these modules to be lightweight, with real-time consideration in mind. Future prediction is supervised using the features of the large model in the current time step, but only during training, so the inference-time speed remains unaffected.

We evaluate the proposed ETA framework on the challenging Bench2Drive benchmark in the \vtwo setting. Our extensive experiments demonstrate that using the large model asynchronously with the smaller model effectively balances performance and latency (\figref{fig:ds_vs_fps}). The proposed model demonstrates significant improvements over baselines, achieving a Driving Score (DS) of 69.53, which represents an improvement of 8.2\% over the previous best DS, particularly in complex scenarios such as merging, overtaking, emergency braking, and yielding. Notably, the dual model achieves these results with a latency of 50 ms, the second lowest on the benchmark, following an MLP-based model~\cite{Zhai2023ARXIV}, which was used as a sanity check.

The contributions are summarized as follows: 
\begin{enumerate}
    \item We introduce a new ETA paradigm to benefit from \lm with real-time consideration;
    \item We propose a dual framework with batched predictive large model inference and timely small model adjustment to implement the ETA paradigm;
    \item We achieve competitive results with (close to) real-time performance in the closed-loop setting on the challenging Bench2Drive benchmark. 
\end{enumerate}

\section{Related Work}
\label{sec:rw}

\subsection{Dual Approach to Embodied Systems}
\boldparagraph{Autonomous Driving} Due to the large computational cost of LLMs and VLMs, a dual approach similar to ours has recently been explored for driving. DriveVLM-Dual~\cite{Tian2024CoRL} uses a larger VLM running asynchronously with a smaller traditional modular pipeline for 3D perception and motion planning by refining coarse waypoints predicted by the VLM. Despite being heavily optimized, 2B-parameter VLM takes 300ms to run, rendering closed-loop evaluation infeasible. 
Given observations and high-level textual instructions, AD-H~\cite{Zhang2024ARXIV} uses an LLM to output a higher level plan in the form of mid-level driving commands, such as turn left, \etc, and conditioned on that, a small LLM predicts low-level waypoints.  
\begin{figure}[t]
    \centering
    \includegraphics[width=0.9\linewidth]{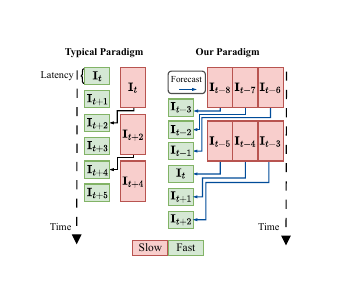}
    \caption{\textbf{Comparison of Dual Approaches.} Typically, dual-system approaches accomodate the larger latency of the slow, low-frequency model (red) by using the slow model predictions a fraction of the time. The fast, high frequency model (green) outputs predictions at every time step, incorporating the slow model's outputs when available. In our approach, we batch and forecast in order to benefit from the larger model at every time step.}
    \label{fig:dual_comp}
\end{figure}
LeapAD~\cite{Mei2024NeurIPS} proposes a dual approach with a heuristic system for fast decisions based on samples retrieved from a memory and a slow analytical system that uses GPT4 for reasoning in the case of failures of the fast part and then stores the corrected action in the memory. %

In a dual paradigm, the frequency of querying large model \vs small model is an important design decision, resulting in a trade-off between accuracy and speed. Our work distinguishes from prior works in that we enable utilization of relevant large model features at every time step.
We propose an efficient dual framework with batched predictive large model inference and timely small model adjustment. %

\boldparagraph{Robotics} LLMs and VLMs are significantly welcomed in the robotics domain to provide open-world reasoning~\cite{Driess2023ICML, Zitkovich2023CoRL}. Very recent works resort to a hierarchical structure to combine the large foundation models and low-level policy networks as a dual framework to realize both instruction decomposition and real-time control~\cite{Zhang2024CoRL, Shentu2024ARXIV}. Li \textit{et al.}~\cite{Li2024ARXIV} propose to encode observations to latents with a VLM in low frequency and employ an action decoder conditioned on the latents with high prediction frequency. Instead, RoboDual~\cite{Bu2024ARXIV} and Hi Robot~\cite{Shi2025ARXIV} integrate different models for the dual system, adopting vision-language-action models for System-2 and System-1, respectively. This methodology has also been successfully applied in industrial companies such as Figure AI~\cite{Figure2025}, where generalized performance and efficient deployment are achieved for whole upper-body control of humanoid robots.

\begin{figure*}[t]
    \centering
    \includegraphics[width=.8\linewidth]{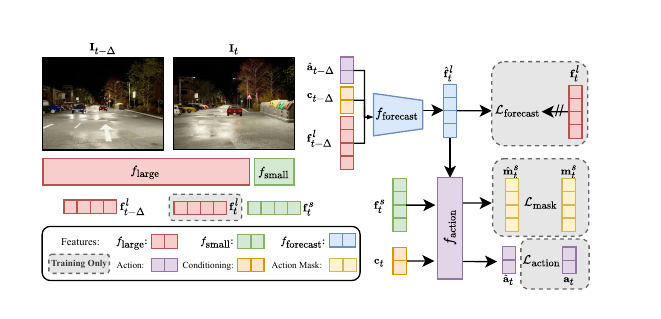}
    \caption{\textbf{Overview of the Asynchronous (Async) Model.} Our model processes two frames, $\Delta$ apart in time, using the large model $f_{\text{large}}$ for the previous frame $\bI_{t-\Delta}$ and the small model $f_{\text{small}}$ for the current frame $\bI_t$. Based on the previous frame's features from the large model, $\bff_{t-\Delta}^l$, along with conditioning inputs $\bc_t$ and $\hat{\ba}_t$, we predict the current time-step features, $\hat{\bff}^l_t$. The action $\hat{\ba}_t$ is then predicted using the action model $f_{\text{action}}$, which takes as input the forecasted features $\hat{\bff}^l_t$, the current frame's features from the small model $\bff_t^s$, and the conditioning input $\bc_t$. The forecasted features are supervised using the large model’s features at the current time step, $\bff_t^l$. Since this supervision occurs during training alone, inference-time speed remains unaffected. Training-time computations are highlighted with gray boxes.
    In addition to forecasting and action losses, we apply a mask loss to better align observations with actions.
    }
    \label{fig:overview}
\end{figure*}
\subsection{Recent Progress on CARLA}
\boldparagraph{Closed-Loop Large Models on Leaderboard-v1} 
Earlier autonomous driving frameworks utilizing \lm evaluate on Town05~\cite{Wang2023ARXIV, Mei2024NeurIPS} or LAV benchmark~\cite{Zhang2024CVPR} in \vone setting for closed-loop evaluation. 
These approaches first process the input using a large encoder, such as Qformer~\cite{Wang2023ARXIV} or CLIP ViT~\cite{Zhang2024CVPR}, to tokenize the scene or a VLM to create a textual description of the scene~\cite{Mei2024NeurIPS} and then feed its output to an LLM/MLLM to predict action.
LangAuto, another benchmark used by LLM/VLM-based agents~\cite{Shao2024CVPR, Zhang2024ARXIV}, provides a semi-closed-loop evaluation but with predefined routes with explicit instructions as input instead of route waypoints. 

\boldparagraph{Leaderboard-v2}
Due to the challenging long scenarios in CARLA \vtwo, there are only a few submissions, each scoring extremely low driving scores. 
The modular~\cite{Zhang2024ARXIV} or hybrid~\cite{Kaljavesi2024ARXIV} approaches fall behind end-to-end approaches. 
End-to-end imitation learning approaches require high-quality driving demonstrations from an expert. 
There are two experts, \ie, with access to privileged data, commonly used for data collection in the \vtwo setting: Think2Drive~\cite{Li2024ECCV}, which is a model-based RL agent, and PDMLite~\cite{Beißwenger2024, Sima2024ECCV}, which is a rule-based planner.
TransFuser++~\cite{Jaeger2023ICCV}, the best imitation learning agent in \vone, performs the second best in \vtwo using data collected by the PDMLite expert. 

LLM4AD/CarLLaVA~\cite{Renz2024ARXIV}, the winner of the CARLA Autonomous Driving Challenge 2.0, encodes the input with LLaVA-Next and then feeds its output to an LLM to predict action. Our base model is similar to LLM4AD/CarLLaVA but is still far behind real-time constraints at 102 ms, which is improved to 50 ms with our dual approach. %
As of December 2024, the CARLA leaderboard test server is temporarily closed and does not accept submissions. 

\boldparagraph{Bench2Drive Benchmark} Due to the difficulty of comparing methods without significant differences in driving scores on challenging, long test routes of Leaderboard-v2, Bench2Drive~\cite{Jia2024NeurIPS} creates a benchmark with short routes, each focused on evaluating a distinct skill set. Several previous end-to-end driving methods are benchmarked by training on official training data collected by the Think2Drive expert~\cite{Li2024ECCV}. 
DriveTransformer~\cite{Jia2025ICLR} proposes a transformer framework with different types of attention, achieving SOTA performance on Bench2Drive; however, it still has a high latency. Our dual framework achieves the best results with significantly lower latency (\figref{fig:ds_vs_fps}).  
Recent work~\cite{Zimmerlin2024ARXIV, Arasteh2024ARXIV} 
reports results by collecting data from other experts, creating an unfair comparison due to changes to training data in terms of resolution, diversity, \etc.

\section{Methodology}
\label{sec:method}

\subsection{Overall}
In this work, we propose a dual system that leverages the concept of shifting intensive computations of large models from the current frame to previous frames. Additionally, we introduce the asynchronous batch inference technique, which performs feature extraction using large models across multiple frames simultaneously, thereby trading space complexity for time complexity. The integration of these two approaches enables timely inference from large models for every frame, effectively unleashing their potential in online systems.

\boldparagraph{Predictive Large Model}
The predictive large model is specifically designed to transfer the computation of the current frame to a previous time step. Given two RGB images, $\bI_t$ and $\bI_{t-\Delta}$, captured $\Delta$ seconds apart, along with the current speed $\bv_t$ and the target waypoints $\bw_t$, our objective is to predict the corresponding driving action $\ba_t$. This action comprises a path made up of equidistant points and varying distance waypoints, as in \cite{Renz2024ARXIV}.

To accomplish this, we first introduce a base model that employs a large encoder (\secref{sec:method_lm}), which, while accurate, is inefficient in terms of processing time. Subsequently, we develop our dual framework (\secref{sec:method_async}), which processes the two time steps asynchronously. This method not only reduces
latency but also retains the advantages of utilizing a large model.

\boldparagraph{Asynchronous Batch Inference}
The concept of asynchronous batch inference is illustrated in \cref{fig:dual_comp}. As demonstrated, to facilitate the use of large models for every frame, we parallelize the inference of previous frames to enable large models for per frame. By simultaneously processing multiple frames, we can significantly reduce the latency burden typically associated with large models, as multiple frames are processed at the same time, enabling more responsive and efficient online systems. This approach not only optimizes resource utilization but also enhances the overall performance of the predictive large model, ensuring timely and accurate driving actions.

\subsection{Base Large Model} 
\label{sec:method_lm}
By using a large model in all time steps, we can train a base model to perform well but with high latency. This allows us to explore the upper bounds of driving performance with various large models if latency were not a concern.  

For example, following the approach proposed by the winner~\cite{Renz2024ARXIV} of the CARLA Autonomous Driving Challenge 2.0, we can use a large vision encoder $f_{\text{large}}$, \eg a VLM or MLLM, to extract features from sensory input $\bI_t$ at time $t$, resulting in $\bff_t^l$. 
We then feed it to an action model $f_{\text{action}}$ to predict action $\hat{\ba}_t$ conditioned on $\bc_{t}$, which is the concatenation of speed $v_t$ and target waypoints $\bw_t$ at time $t$:
\begin{eqnarray}
    \bff_t^l &=& f_{\text{large}}(\bI_t), \\
    \hat{\ba}_t &=& f_{\text{action}}(\bI_t, \bc_t).
\end{eqnarray}
Following LLaVA's anyres \cite{Liu2024CVPR,Liu2023NeurIPS} approach, we divide the image into two patches and then process them with the vision encoder. To reduce the number of output tokens, we apply $2\times2$ spatial pooling, decreasing the token count by a factor of 4. Despite this reduction, the best-performing base models with a large encoder still fail to achieve a latency close to real-time constraints.

\begin{figure}[t]
    \centering
    \includegraphics[width=\linewidth]{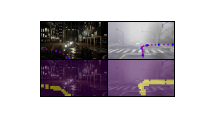}
    \caption{\textbf{Action Mask.} In the top row, we show two examples of the RGB image with the path (purple) and waypoints (blue) projected onto it. Patches containing a path or waypoint are marked as 1 (yellow), and all other patches are marked as 0 (purple), creating the binary mask, overlayed on the image below for visualization.}
    \label{fig:action_mask}
\end{figure}

\boldparagraph{Action Mask} 
When we visualized the attention maps, we observed that the model struggles to focus on image patches that are relevant to action. To better align the predicted actions with the observations, we compute attention between patch features and the encoded queries corresponding to the action. This process generates a mask, $\hat{\bm}_t$, showing the regions of the input image toward which the ego vehicle is likely to move (\figref{fig:action_mask}). We supervise this action mask with the ground truth mask, which is obtained by projecting expert actions onto the image (\secref{sec:training}).

\subsection{Asynchronous (Async) Model}
\label{sec:method_async}
As illustrated in \figref{fig:overview}, we propose a dual framework with a slow, large model $f_{\text{large}}$ and a smaller, faster model $f_{\text{small}}$. 
Our small model is designed with real-time performance in mind. However, large models with strong performance typically exceed real-time constraints. 
To utilize the large model without increasing latency, we process the input at a previous time step $t-\Delta$, rather than the current time step $t$:
\begin{equation}
    \bff_{t-\Delta}^l = f_{\text{large}}(\bI_{t-\Delta}).
\end{equation}
$\bI_{t-\Delta}$ denotes the input image at time $t-\Delta$ with the corresponding features $\bff_{t-\Delta}^l$ obtained from the large model. 

\boldparagraph{Forecasting} Given features $\bff_{t-\Delta}^l$ from the large model at a previous time step, we train a model $f_{\text{forecast}}$ to predict features at the current time step. Similar to a world model~\cite{Li2025ICLR}, we condition this model on additional inputs, including the action predicted, $\hat{\ba}_{t-\Delta}$, and $\bc_{t-\Delta}$, which is the concatenation of speed $v_{t-\Delta}$ and target waypoints $\bw_{t-\Delta}$: %
\begin{equation}
   \hat{\bff}^l_t = f_{\text{forecast}}(\bff_{t-\Delta}^l, \hat{\ba}_{t-\Delta}, \bc_{t-\Delta}). 
   \label{eq:forecast}
\end{equation}
The result of forecasting is the predicted features $\hat{\bff}^l_t$ for the current time step $t$.

\boldparagraph{Small Model} In addition to forecasting the features for the current time step, we use the small model, $f_{\text{small}}$, to process the current input $\bI_t$, resulting in features $\bff_t^s$:
\begin{equation}
    \bff_t^s = f_{\text{small}}(\bI_t).
\end{equation}
The small model is designed to capture changes between the previous and current time steps, particularly those that are difficult to predict, such as a traffic light’s state or the sudden appearance of a pedestrian.

\boldparagraph{Action Prediction}
We concatenate the predicted features $\hat{\bff}_t$ from forecasting with the features $\bff_t^s$ from the small model and provide them as input to the action model, $f_{\text{action}}$, along with the conditioning input $\bc_t$, resulting in the predicted action $\hat{\ba}_t$:
\begin{equation}
    \hat{\ba}_t = f_{\text{action}}(\hat{\bff}^l_t, \bff_t^s, \bc_t).
\end{equation}

\noindent
It is observed 
that the key to performance lies in forecasting and the small model working together to efficiently update past information from the large model to the current state.

\subsection{Training}
\label{sec:training}
\boldparagraph{Action Loss} For the prediction of action, we apply an $L_1$ loss between the predicted actions and the expert action, including the path and waypoints. Instead of directly predicting absolute positions, we learn to predict residuals. At test time, we sum these residuals to reconstruct the original waypoints. A similar approach is applied to the path tokens. We find that this formulation stabilizes training compared to directly regressing path and waypoint coordinates.  

\boldparagraph{Action Mask Loss} To generate supervision for the action mask, we project the expert action onto the input RGB image using the camera parameters. Patches containing a waypoint or path point are assigned a value of 1, while all others are set to 0 (\figref{fig:action_mask}). We apply a binary cross-entropy loss between the ground truth mask $\bm_t$ created this way and the predicted mask $\hat{\bm}_t$.

The final loss of the base model is a weighted sum of the two loss terms where $\lambda_1 = 1/16$: 
\begin{align}
    \cL_{\text{base}} &= \cL_{\text{action}} + \lambda_1~\cL_{\text{mask}}.
    \label{eq:loss_base}
\end{align}

\boldparagraph{Async Model with Forecasting Loss} For the async model, we optimize the predicted action loss and mask loss as in the case of the base model \eqref{eq:loss_base}, but the mask loss is applied to the features of the small model in the async case.
Additionally, we introduce a forecasting loss, $\cL_{\text{forecast}}$, to supervise the predicted features based on the large model's features from the previous time step. With the forecasting loss, we aim to bring the predicted features closer to the ground truth features extracted from the current time step using the same large model. Specifically, we compute the absolute difference between ground truth features $\bff_t^l$ and the predicted features $\hat{\bff}_t^l$ from \eqref{eq:forecast}:
\begin{align}
        \cL_{\text{forecast}}(\bff_t^l, \hat{\bff}_t^l) = \lvert \bff_t^l - \hat{\bff}_t^l \rvert.
\end{align}
We stop gradients from flowing through the ground truth features to prevent collapse. 
The weight of the mask loss remains the same as in the base model, and we set the weight of the forecasting loss to 0.5.
\section{Experiments}
\label{sec:exp}
\begin{table*}[t!]
\centering
\small
\caption{\textbf{Comparison to SOTA on Bench2Drive.} In addition to the existent baselines,
we compare our method to the recent DriveTransformer~\cite{Jia2025ICLR}, selecting the best-performing variant for each model. Both our base and async models outperform \textit{all} prior arts, with the async model achieving significantly lower latency with the proposed dual framework. Latency is reported in milliseconds. 
}
\begin{tabular}{l r r r r r}
\toprule
\textbf{Method} & \textbf{DS$\uparrow$} & \textbf{SR$\uparrow$(\%)} & \textbf{Efficiency$\uparrow$} & \textbf{Comfort$\uparrow$} & \textbf{Latency$\downarrow$} \\
\midrule
    AD-MLP~\cite{Zhai2023ARXIV} & 18.05 & 0.00 & 48.45 & 22.63 & 3 \\
    UniAD-Base~\cite{Hu2023CVPR} & 45.81 & 16.36 & 129.21 & 43.58 & 663.4 \\
    VAD~\cite{Jiang2023ICCV} & 42.35 & 15.00 & 157.94 & 46.01 & 278.3 \\
    TCP-traj~\cite{Wu2022NeurIPS} & 59.90 & 30.00 & 76.54 & 18.08 & 83 \\
    ThinkTwice~\cite{Jia2023CVPR} & 62.44 & 31.23 & 69.33 & 16.22 & 762 \\
    DriveTransformer-Large~\cite{Jia2025ICLR} & 63.46 & 35.01 & 100.64 & 20.78 & 211.7 \\
    DriveAdapter~\cite{Jia2023ICCV} & 64.22 & 33.08 & 70.22 & 16.01 & 931 \\
    \midrule
    Base Model (Ours) & 74.33{\small$\pm$0.99} & 48.33{\small$\pm$3.56} & 186.04{\small$\pm$6.47} & 25.77{\small$\pm$1.13} & 102 \\
    Async Model (Ours) & 69.53{\small$\pm$1.62} & 38.64{\small$\pm$0.98} & 184.51{\small$\pm$1.54} & 28.43{\small$\pm$1.50} & 50 \\
\bottomrule
\end{tabular}
\label{tab:sota}
\end{table*}

\subsection{Training Setting}
We train all models for 40 epochs on 8$\times$ A100 GPUs. We use CLIP-ViT-L-336px as the base large model encoder and the first 8 layers of CLIP-ViT-L-336px as the small model encoder, which we ablate in Supplementary. For the action decoder, we use a 12-layer transformer decoder.  We use the base Adam optimizer with an initial learning rate of $3 \cdot 10^{-5}$ and a cosine annealing schedule. The learning rate (LR) resets 4 times before reaching the minimum LR near the end of training. We set the global batch size to 160. For the Async Model, we set $\Delta$ to 0.5s.

\subsection{Evaluation Details}
\boldparagraph{Data} The Bench2Drive \cite{Jia2024NeurIPS} Benchmark comes with two data splits collected using the Think2Drive RL agent~\cite{Li2024ECCV}. The base split, which consists of 1000 routes, is used to train and evaluate multiple models in the original paper and following work. To maintain direct comparability to the evaluated models, we only train on the base split and do not use any additional data. We divide the data into a 94\%-3\%-3\% training-validation-test split during our experiments. Similarly to CarLLaVA \cite{Renz2024ARXIV}, we create different buckets containing different interesting scenarios, such as turning, near-collision, or intersections. During training, we use a weighted sampling approach to sample from the buckets. Please see Supplementary for details.
\boldparagraph{Metrics} We report the Driving Score (DS) and Success Rate (SR), along with Comfortness and Efficiency scores. Additionally, we provide per-ability scores to assess the model’s performance across different scenario categories. Finally, we include the latency of each model in milliseconds (ms). For our main models, we report results averaged over three runs with different seeds to account for variance in initialization and closed-loop evaluation. However, due to time and computational constraints, we report results from a single seed for ablation studies.

\begin{table*}[t!]
\centering
\caption{\textbf{Ablation Study.} We conduct an ablation study by removing each component of the model (\textbf{A–C}). Additionally, we report the performance of the small model alone (\textbf{D}) and further analyze forecasting by using ground truth features during both training and testing (\textbf{E}) or only during testing (\textbf{F}). 
`GT' indicates ground truth.
See the text for a detailed analysis.
}
\small
\begin{tabular}{l l r r r r r }
\toprule
\textbf{ID} & \textbf{Method} & \textbf{DS$\uparrow$} & \textbf{SR$\uparrow$(\%)} & \textbf{Efficiency$\uparrow$} & \textbf{Comfort$\uparrow$} & \textbf{Latency$\downarrow$} \\
\midrule
    & Base Model (Ours) & 74.33 & 48.33 & 186.04 & 25.77 & 102 \\
    & Async Model (Ours) & 69.53 & 38.64 & 184.51 & 28.43 & 50 \\
    \midrule
    A & Without Forecasting & 54.92 & 30.45 & 177.52 & 18.71 & 50 \\
    B & Without Small Model & 42.49 & 17.27 & 170.94 & 14.10 & 31 \\
    C & Without Action Mask & 42.67 & 17.27 & 167.32 & 27.99 & 50 \\
    \midrule
    D & Small Model Only & 61.30 & 35.00 & 183.63 & 43.91 & 50 \\
    E & GT Forecasting & 74.12 & 48.18 & 192.72 & 24.76 & 124 \\
    F & GT Forecasting (Test Time) & 60.72 & 32.27 & 151.53 & 17.25 & 124 \\
\bottomrule
\end{tabular}

\label{tab:abl}
\end{table*}
\begin{table*}[t!]
\centering
\caption{\textbf{Ablation Study according to Distinct Abilities.} To identify which components contribute to improvements in different scenarios, we report model performance in the ablation study based on distinct capabilities. For instance, removing the small model (\textbf{C}) leads to significant performance drops in the emergency brake (\textbf{E. Brake}) and traffic sign (\textbf{T. Sign}). See the text for a detailed analysis.}
\small
\begin{tabular}{l l r r r r r r r r}
\toprule
\textbf{ID} & \textbf{Method} & \textbf{Merging$\uparrow$} & \textbf{Overtaking$\uparrow$} & \textbf{E. Brake$\uparrow$} & \textbf{Give Way$\uparrow$} & \textbf{T. Sign$\uparrow$} & \textbf{Mean$\uparrow$} & \textbf{Latency$\downarrow$} \\
\midrule
    & Base Model (Ours) & 35.24%
    & 54.70 & %
    66.67 & %
    80.00 & %
    59.43 & %
    59.21 %
    & 102 \\
    & Async Model (Ours) & 26.66 %
    & 50.42 & %
    60.13 & %
    80.00 & %
    43.64 & %
    52.17 & %
    50 \\
    \midrule
    A & Without Forecasting & 20.00 & 35.90 & 41.18 & 80.00 & 36.18 & 42.65 & 50 \\
    B & Without Small Model & 14.86 & 17.50 & 16.36 & 40.00 & 30.56 & 23.85 & 31 \\
    C & Without Action Mask & 20.00 & 17.95 & 9.80 & 80.00 & 30.92 & 31.73 & 50 \\
    \midrule
    D & Small Model Only & 27.14 & 41.03 & 47.06 & 60.00 & 44.74 & 43.99 & 50 \\
    E & GT Forecasting & 29.33 & 57.50 & 56.36 & 60.00 & 53.89 & 51.41 & 124 \\
    F & GT Forecasting (Test) & 28.57 & 38.46 & 43.14 & 60.00 & 43.42 & 42.72 &  124 \\
\bottomrule
\end{tabular}
\label{tab:abl_sep}
\end{table*}

\subsection{Comparison to State-of-the-Art}
In \tabref{tab:sota}, we compare 
our method 
to other approaches,
including the originally reported AD-MLP~\cite{Zhai2023ARXIV}, UniAD~\cite{Hu2023CVPR}, VAD~\cite{Jiang2023ICCV}, TCP~\cite{Wu2022NeurIPS}, ThinkTwice~\cite{Jia2023CVPR}, and DriveAdapter~\cite{Jia2023ICCV}. 
Note that we are unable to quantitatively compare
to 
other 
dual approaches \cite{Mei2024NeurIPS,Tian2024CoRL,Zhang2024ARXIV} due to the lack of comparable closed-loop results.

Our base model, without asynchronous future prediction, achieves the highest driving score among all counterparts,
with a latency of 102 ms. %
However, despite the strong performance of our base model, its 102ms latency imposes an upper bound of approximately 10Hz on the control frequency, which is relatively low for real-time control.

Using our asynchronous model, which performs the computation of the full vision encoder asynchronously until the time it is needed, we reduce latency to 50ms, enabling a control frequency of 20Hz.  This marks a significant improvement over the base model, allowing for much more responsive control. 
Despite not having access to the large model at the current time step, our async model still performs very well in terms of driving score and success rate, ranking second after our base model. %

\subsection{Ablation Studies}

We perform an ablation study of our design choices in \tabref{tab:abl} to address the following research questions:
\begin{itemize} 
    \item Does supervising the large model for forecasting improve performance, or is the improvement solely due to scaling up the model parameters? 
    \item Is the small model necessary, or can we rely solely on the forecasted features? 
    \item Is additional supervision with the action mask essential?
    \item How does the small model perform without forecasting? 
    \item Would better forecasting lead to better performance? 
    \item In which scenarios do we observe improvements, and what are the underlying reasons? 
\end{itemize}

\boldparagraph{A} \textbf{Without supervising forecasting}, the driving score drops to 54.92, highlighting the importance of training the large model to learn forecasting. The performance improvement cannot be solely attributed to the increased parameter count, as the large model’s features do not enhance performance without being explicitly trained for future prediction. By supervising the model to forecast with ground truth features, the large vision encoder effectively improves the driving score by approximately 15 points. 
Interestingly, the overall performance remains lower than when using only the small model (\textbf{D}), possibly due to confusion caused by suboptimal features from the large model.

\boldparagraph{B} \textbf{Without small model}, the performance drops to 42.49,  indicating that forecasting with the large model alone is insufficient without up-to-date information from the small model. 
This may be due to scenarios where the future state is difficult to predict, such as changes in traffic light status. As shown in \tabref{tab:abl_sep}, removing the small model (\textbf{B}) results in a significant decline in emergency braking ability (60.13$\rightarrow$16.36) and proper handling of traffic signs (43.64$\rightarrow$30.56). These results highlight the critical role of the small model in providing the model with up-to-date information by allowing it to glance at the current state.

\begin{figure*}[t]
    \centering
    \includegraphics[width=\linewidth]{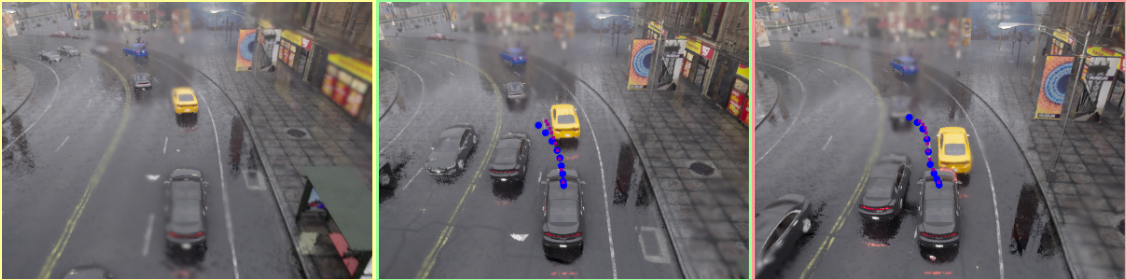}
    \caption{\textbf{Hard Brake without Small Model.} \textbf{Left:} The initial state of the scene where the yellow car ahead suddenly brakes, requiring the ego vehicle to perform a hard brake. \textbf{Middle:} The Async Model detects the hazard using the small model and stops in time. \textbf{Right:} The version without the small model (\textbf{B}) relies solely on forecasting and fails to detect the sudden stop in time, resulting in a crash. The predicted action for each model is overlaid on the image, with waypoints shown in blue and the path in red.}
    \label{fig:hard_brake}
\end{figure*}

\begin{figure*}[t]
    \centering
    \includegraphics[width=\linewidth]{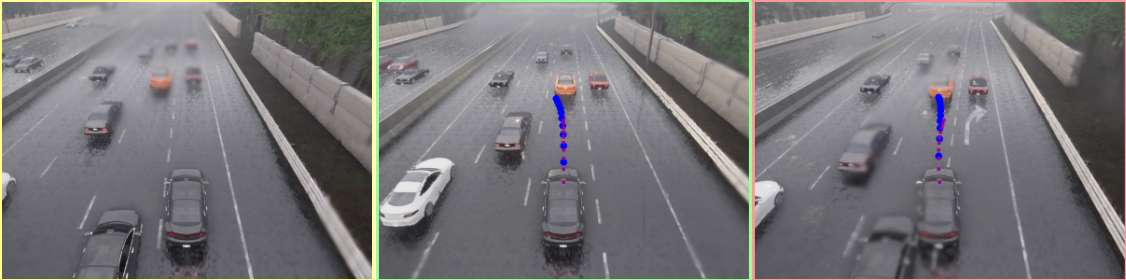}
    \caption{\textbf{Lane Change without Forecasting.} \textbf{Left:} The initial state of the scene, where the ego vehicle is tasked with switching to the left lane. \textbf{Middle:} The Async Model successfully executes the lane change while avoiding collisions by leveraging forecasting. \textbf{Right:} The version without forecasting (\textbf{A}) fails to anticipate the future position of the vehicle behind, resulting in a rear-end collision. The predicted action for each model is overlaid on the image, with waypoints shown in blue and the path in red.}
    \label{fig:lane_change}
\end{figure*}

\boldparagraph{C} \textbf{Without action mask,} the performance drops to 42.67, highlighting the importance of aligning the predicted actions with the observations. It might be surprising that such a small adjustment has such a significant impact on the async model, but this was one of the first insights we gained while attempting to make the base model work by examining the visualization of attention maps. This additional supervision helps the model better relate the predicted action to the observations. We found that removing it had a less severe effect on the base model (see Supplementary).

\boldparagraph{D} \textbf{Using only the small model}, performance drops to 61.30, demonstrating that the predicted features from the large model enhance overall performance. With our dual formulation, we can leverage these features without impacting latency. 
Comparing the async model with forecasting to using only the small model without forecasting (\textbf{D}) in terms of abilities in \tabref{tab:abl_sep}, we observe a notable decline in overtaking (50.42$\rightarrow$41.03) and emergency braking (60.13$\rightarrow$47.06) behaviors. 
A possible explanation is that the large model provides strong features for predictable scenarios, enabling the small model to focus on detecting deviations, such as emergencies. Without the large model, the small model must handle all situations, leaving fewer resources for effectively managing emergencies.

\boldparagraph{E \& F} \textbf{Using ground truth features for forecasting}, we further examine the impact of forecasting on performance. 
When ground truth features replace forecasted features during both training and testing (\textbf{E}), the driving score improves to 74.12, surpassing the score achieved with our async model's forecasted features (69.53). This gap suggests that forecasting can still be improved, and with perfect forecasting, the model can perform at the same level as our base model (74.33), ruling out potential design limitations. 
However, when ground truth features are provided only at test time (\textbf{F}), the driving score drops significantly to 60.72. %
This suggests a distribution mismatch between the predicted and ground truth features, even though the model is trained with supervision from ground truth features.

\subsection{Qualitative Analysis}
We visually investigate the contributions of the small model and the large model with forecasting through two example scenarios: a hard brake without the small model (\figref{fig:hard_brake}) and a lane change without forecasting supervision (\figref{fig:lane_change}). These correspond to rows \textbf{B} and \textbf{A} in the ablation study (\tabref{tab:abl}), respectively. In \figref{fig:hard_brake}, the version in \textbf{B} fails to perform a hard brake, as it cannot detect the sudden stop of the yellow vehicle ahead without the small model, highlighting its role in detecting emergencies that appear in the current time step. Similarly, in \figref{fig:lane_change}, the version in \textbf{C} results in a rear-end crash due to its failure to anticipate the future position of the vehicle behind without forecasting.

\section{Conclusion}
\label{sec:conc}

We introduce a novel dual paradigm that shifts computation to the previous time step, enabling more efficient processing with large models in potential applications in embodied systems. To implement this paradigm in self-driving, we proposed ETA, a dual framework that combines forecasting to propagate valuable information from a large model at the prior time step to the current time step and a small model to capture unpredictable changes, such as sudden brakes due to emergencies. ETA advances state-of-the-art driving performance on Bench2Drive by 8\%, also reducing latency from hundreds of milliseconds in previous methods to just 50 ms. Through detailed ablation studies, we demonstrate the importance of small model and forecasting based on the large model's output, highlighting the effectiveness of ETA to combine the two in a dual framework. 

\boldparagraph{Limitations and Future Work}
While our async model significantly narrows the gap to the base model in critical driving scenarios such as overtaking, emergency braking, and yielding, some challenges remain. The largest discrepancies between the base and async models occur in the traffic sign handling and merging scenarios. 

Future work will explore improving forecasting, as our analysis shows that using ground truth features instead of forecasting brings the async model to the level of the base model. Furthermore, we see potential for pushing performance even further by leveraging larger and more advanced vision encoders, which could enhance both perception and decision-making capabilities.

\section*{Acknowledgements}
This project is funded by the European Union (ERC, ENSURE, 101116486) with additional compute support from Leonardo Booster (EuroHPC Joint Undertaking, EHPC-AI-2024A01-060). Views and opinions expressed are however those of the author(s) only and do not necessarily reflect those of the European Union or the European Research Council. Neither the European Union nor the granting authority can be held responsible for them.
This study is also supported by National Natural Science Foundation of China (62206172) and Shanghai Committee of Science and Technology (23YF1462000).
We extend our gratitude to Li Chen and the rest of the members from OpenDriveLab for their profound discussions and supportive help in writing and related work. We also extend our gratitude to Görkay Aydemir and the rest of the members from the Autonomous Vision Group at Koc University for their constructive feedback and discussions. 

{
    \small
    \bibliographystyle{ieeenat_fullname}
    \bibliography{main}
}
\newpage 
\clearpage
\begin{abstract}
    In this supplementary material, we first review related work on using LLMs/VLMs in driving (\secref{sec:add_rw}). Next, we provide experimental details to ensure reproducibility (\secref{sec:exp_details}). Finally, we present additional quantitative and qualitative results in \secref{sec:add_results}, including the comparison of models according to abilities 
    and an exploration of alternative models by varying the large encoder in the base model.
\end{abstract}

\section{Additional Related Work}
\label{sec:add_rw}
\boldparagraph{LLMs/VLMs for Driving} There is a large increase in the use of LLMs/VLMs for driving. A line of work, represented by approaches like DriveMLM~\cite{Wang2023ARXIV}, LLM4AD/CarLLaVA~\cite{Renz2024ARXIV}, and FeD~\cite{Zhang2024CVPR}, uses LLMs for initialization to benefit from their pre-training on large datasets. A typical approach is to encode the scene using smaller models and feed the encoded visual features into the LLM to predict action. For fast inference, especially in closed-loop evaluation, these approaches prefer small language models, \eg, as small as 50M in \cite{Renz2024ARXIV}. In this line of work, LLMs are not fully utilized, \ie, for their reasoning capabilities. As reported in \cite{Renz2024ARXIV}, the improvement mostly comes from the large-scale pre-training of the vision encoder rather than the LLM on top. 

Another line of work, including approaches like DriveVLM~\cite{Tian2024CoRL} and DriveLM~\cite{Sima2024ECCV}, aims to benefit from the reasoning capabilities of LLMs, \eg with Chain-of-thought (CoT) reasoning. This kind of approach has the potential to fully utilize the power of LLMs for driving, \eg by improving out-of-distribution cases with their pre-trained knowledge. However, multi-step reasoning causes high latency, restricting these models to open-loop evaluation. Similarly, our goal is to benefit from the pre-training knowledge of LLMs as much as possible, but differently, without sacrificing real-time performance in closed-loop evaluation.

\boldparagraph{Other Uses of LLMs for Driving} Besides action, previous work also uses an LLM to detect hard cases in motion prediction~\cite{Yang2024IV} or to align features of a driving model with the features of an LLM~\cite{Pan2024CVPR}. Another work analyzes the capabilities of LLMs in terms of future prediction~\cite{Sreeram2024ARXIV}. 

\section{Experimental Details}
\label{sec:exp_details}

\boldparagraph{Base Large Model} For the Large Model, we use the CLIP-ViT-L-336px encoder from the checkpoint of LLaVA 1.6 \cite{Liu2024CVPR} 7B Vicuna, which has 24 layers with a hidden dimension size of 1024 and an input patch size of 14.

\boldparagraph{Small Model} For the Small Model, we use the first 8 layers of the same CLIP-ViT-L-336px encoder as the Base Large Model.

\boldparagraph{Action Prediction Model} For the Action Prediction Model, we use the autoregressive Llama architecture initialized from scratch, with 12 hidden layers and a hidden size of 768.

\boldparagraph{Forecasting Model} For the Forecasting Model, we use a lightweight transformer encoder with 2 layers. We concatenate the input features with the conditioning along the feature dimension, project first to the feature dimension size, then pass to the transformer encoder.

\boldparagraph{Training Setting} We use the Deepspeed zero2 setting during training for memory optimization. We use bfloat16 during training, and do not use any weight decay.

\begin{table*}[t!]
\centering
\caption{\textbf{Comparison to SOTA according to Distinct Abilities.} We compare our async model to state-of-the-art methods across distinct abilities, highlighting significant improvements in all challenging scenarios except traffic sign (T. Sign) over previous best results.}
\small
\begin{tabular}{l r r r r r r r }
\toprule
\textbf{Method} & \textbf{Merging$\uparrow$} & \textbf{Overtaking$\uparrow$} & \textbf{E. Brake$\uparrow$} & \textbf{Give Way$\uparrow$} & \textbf{T. Sign$\uparrow$} & \textbf{Mean$\uparrow$} & \textbf{Latency$\downarrow$} \\
\midrule
    AD-MLP~\cite{Zhai2023ARXIV} & 0.00 & 0.00 & 0.00 & 0.00 & 0.00 & 0.00 & 3 \\
    UniAD-Base~\cite{Hu2023CVPR} & 12.16 & 20.00 & 23.64 & 10.00 & 13.89 & 15.94 & 663.4 \\
    VAD~\cite{Jiang2023ICCV} & 7.14 & 20.00 & 16.36 & 20.00 & 20.22 & 16.75 & 278.3 \\
    TCP-traj~\cite{Wu2022NeurIPS} & 12.50 & 22.73 & 52.72 & 40.00 & 46.63 & 34.92 & 83 \\
    ThinkTwice~\cite{Jia2023CVPR} & 13.72 & 22.93 & 52.99 & 50.00 & 47.78 & 37.48 & 762 \\
    DriveTransformer-Large~\cite{Jia2025ICLR} & 17.57 & 35.00 & 48.36 & 40.00 & 52.10 & 38.60 & 211.7 \\
    DriveAdapter~\cite{Jia2023ICCV} & 14.55 & 22.61 & 54.04 & 50.00 & 50.45 & 38.33 & 931 \\
    \midrule
    Base Model (Ours) & 35.24%
    & 54.70 & %
    66.67 & %
    80.00 & %
    59.43 & %
    59.21 %
    & 102 \\
    Async Model (Ours) & 26.66%
    & 50.42 & %
    60.13 & %
    80.00 & %
    43.64 & %
    52.17 %
    & 50 \\
\bottomrule
\end{tabular}
\label{tab:res}
\end{table*}

\boldparagraph{Data Bucketing} To increase the frequency of interesting data samples, we divide samples to buckets as in \cite{Renz2024ARXIV}. Additionally, we weight the buckets to oversample some buckets. The buckets we design are: 

\begin{itemize}
    \item Acceleration from scratch: The agent is currently stationary (speed $<0.05$) and and starting to accelerate.
    \item Light acceleration: The agent is slightly accelerating ($0.2 < \text{throttle} < 0.5$). We weight this bucket with a weight of 2.
    \item Medium acceleration: The agent is slightly accelerating ($0.5 < \text{throttle} < 0.9$). We weight this bucket with a weight of 2.
    \item Strong acceleration: The agent is slightly accelerating ($\text{throttle} > 0.9$).
    \item Braking: The agent is braking.
    \item Coasting: The agent is at a non-zero speed, is not braking, and is not accelerating strongly ($\text{throttle} < 0.2$).
    \item Steering Left: The agent steering to the left. We weight this bucket with a weight of 3.
    \item Steering Right: The agent steering to the right. We weight this bucket with a weight of 3.
    \item Rear Vehicle Hazard: Assuming all agents maintain speed and heading, there is at least one other vehicle that will hit the ego vehicle from behind (rear 60 degrees).
    \item Front Vehicle Hazard: Assuming all agents maintain speed and heading, there is at least one other vehicle that will collide with the ego vehicle from the front (front 60 degrees).
    \item Side Vehicle Hazard: Assuming all agents maintain speed and heading, there is at least one other vehicle that will collide with the ego vehicle from either side (left and right, 120 degrees each).
    \item Stop Sign: The ego vehicle is currently within the range of a stop sign.
    \item Red Light: The ego vehicle is currently within the range of a red traffic light.
    \item Swerving Bucket: The current route contains a scenario which is is one of:
    \begin{itemize}
        \item Accident
        \item BlockedIntersection
        \item ConstructionObstacle
        \item HazardAtSideLane
        \item ParkedObstacle
        \item VehicleOpensDoorTwoWays
    \end{itemize}
    \item Pedestrian: Assuming all pedestrians maintain speed and heading, there is at least one pedestrian that will be hit by the ego vehicle.
\end{itemize}

\section{Additional Experiments}
\label{sec:add_results}

\subsection{Ability Scores}
We include the per-ability scores for all models in \tabref{tab:res}.

\subsection{Additional Ablations}
We include two additional ablations in \tabref{tab:ablsupp} in addition to the ablations inincluded in the main paper. 

\boldparagraph{Action Mask Loss} We ablate the inclusion of the action mask loss \textbf{E} for the Base Model. Compared to the Async Model setting \textbf{C} in the main paper (Table 2), the decrease in the Driving Score is relatively lower.

\boldparagraph{Different Small Model} We experiment with using ViT-Base as the small model (\textbf{E}), despite it not being fast enough to meet real-time constraints. We find it less suitable as the small model, with Driving Score decreasing significantly (57.07 \vs 69.53).

\begin{table}[b!]
\centering
\small
\caption{\textbf{Varying the Large Encoder in the Base Model.} We experiment with various sizes of large encoders in the base model to observe the differences when changing our default large encoder in \textbf{A} to smaller encoders (\textbf{B–E}) and larger encoders (\textbf{F, G}). While larger encoders generally perform better, smaller encoders achieve lower latencies, as expected. Notably, the model in \textbf{B} has a similar latency to our Aysnc model but performs significantly worse (61.30 \vs 69.53). \textbf{Lat.} stands for latency in milliseconds.}
\begin{tabular}{l l r r r r}
\toprule
\textbf{ID} & \textbf{Model} & \textbf{Size} & \textbf{DS$\uparrow$} & \textbf{SR$\uparrow$(\%)} & \textbf{Lat.$\downarrow$} \\
\midrule
    A & CLIP-ViT-L%
    & 308M & 74.33 & 48.94 & 102 \\
    B & CLIP-ViT-L8%
    & 114M & 61.30 & 35.00 & 50 \\
    C & InternViT%
    & 304M & 71.53 & 46.82 & 132 \\ %
    D & ViT-Base%
    & 86.9M & 67.53 & 44.55 & 67 \\
    E& EVA02-Base%
    & 85.8M & 65.20 & 37.73 & 80 \\
    F & CLIP ViT-H%
    & 631M & 70.74 & 43.64 & 144 \\
    G & CLIP-ViT-g
    & 1011M & 73.19 & 45.91 & 192 \\
\bottomrule
\end{tabular}

\label{tab:abl_enc}
\end{table}

\begin{table*}[t!]
\centering
\caption{\textbf{Ablation Study.} We conduct an ablation study by removing each component of the model (\textbf{A–C}). Additionally, we report the performance of the small model alone (\textbf{D}) and further analyze forecasting by using ground truth features during both training and testing (\textbf{E}) or only during testing (\textbf{F}). 
`GT' indicates ground truth.
See the text for a detailed analysis.
}
\small
\begin{tabular}{l l r r r r r }
\toprule
\textbf{ID} & \textbf{Method} & \textbf{DS$\uparrow$} & \textbf{SR$\uparrow$(\%)} & \textbf{Efficiency$\uparrow$} & \textbf{Comfort$\uparrow$} & \textbf{Latency$\downarrow$} \\
\midrule
    & Base Model (Ours) & 74.33 & 48.33 & 186.04 & 25.77 & 102 \\
    & Async Model (Ours) & 69.53 & 38.64 & 184.51 & 28.43 & 50 \\
    \midrule
    A & Without Forecasting & 54.92 & 30.45 & 177.52 & 18.71 & 50 \\
    B & Without Small Model & 42.49 & 17.27 & 170.94 & 14.10 & 31 \\
    C & Base Model Without Action Mask & 70.02 & 87.26 & 178.37 & 29.97 & 102 \\
    D & Async Model Without Action Mask & 42.67 & 17.27 & 167.32 & 27.99 & 50 \\
    E & ViT-Base as Small Model & 57.07 & 27.73 & 175.58 & 35.18 & 67 \\
    \midrule
    F & Small Model Only & 61.30 & 35.00 & 183.63 & 43.91 & 50 \\
    G & GT Forecasting & 74.12 & 48.18 & 192.72 & 24.76 & 124 \\
    H & GT Forecasting (Test Time) & 60.72 & 32.27 & 151.53 & 17.25 & 124 \\
\bottomrule
\end{tabular}

\label{tab:ablsupp}
\end{table*}

\subsection{Additional Qualitative Analysis}
\begin{figure*}
    \centering
    \includegraphics[width=\linewidth]{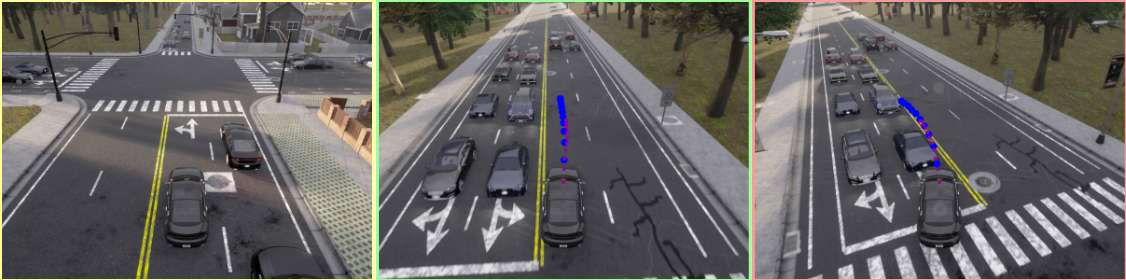}
    \caption{\textbf{Turn Failure.} In the left column, we show the initial state of the scene. In this scenario, the agent is performing a right turn at an intersection. In the middle, the Async Model can maneuver properly, performing a correct turn. On the right, the model without the small encoder (\textbf{B}) is unable to perform a smooth turn, performing a tighter turn resulting in a crash.}
    \label{fig:turn}
\end{figure*}

\begin{figure*}
    \centering
    \includegraphics[width=\linewidth]{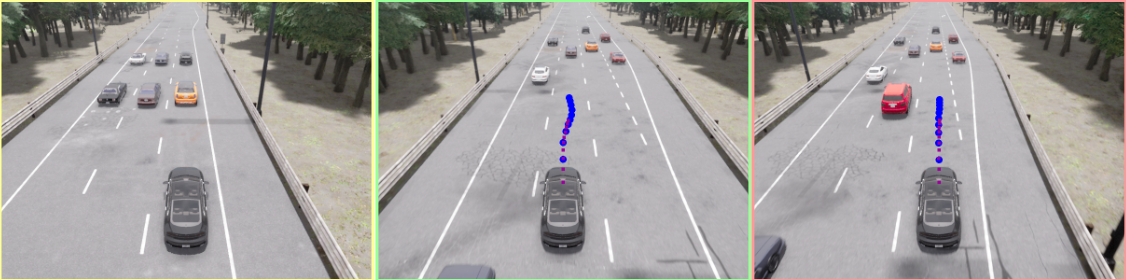}
    \caption{\textbf{Lane Change - Base model \vs Async.} In the left column, we show the initial state of the ego vehicle in traffic. In this scenario, the ego vehicle is required to switch the left lane. In the middle, the Base model is able to correctly make the lange change while avoiding collision. Interestingly, the Async model on the right ignores the requested lane change and continues driving straight. Due to the leaderboard evaluation method, both models get a perfect driving score in this scenario.}
    \label{fig:lanev2}
\end{figure*}

We present two additional qualitative examples: one comparing the Async model to the version without the small model (\textbf{B}) in \figref{fig:turn}, and another examining lane change behavior between the Base model and the Async model in \figref{fig:lanev2}.

\end{document}